\documentclass[11pt]{article}
\usepackage{graphicx}
\usepackage{subfig}
\usepackage{amssymb}
\usepackage{amsmath}
\usepackage{amsthm}
\usepackage{makeidx}
\usepackage{algorithm}
\usepackage[noend]{algorithmic}
\algsetup{indent=2em}
\usepackage{amsfonts}
\usepackage{latexsym}
\usepackage{boites}
\usepackage{url}
\usepackage{placeins}
\usepackage{float}
\usepackage{mathrsfs}
\usepackage{amsbsy}
\usepackage{bm}
\usepackage{fixmath}
\usepackage[square,sort,comma,numbers]{natbib}
\usepackage{psfrag}
\usepackage{xcolor}
\usepackage[T1]{fontenc}
\usepackage[margin=1in]{geometry}
\usepackage[hidelinks]{hyperref}

\usepackage{cleveref}
\usepackage[most]{tcolorbox}
\usepackage{mparhack}
\usepackage{titletoc}
\usepackage{booktabs}
\newtheorem{theorem}{Theorem}[section]
\newtheorem{lemma}[theorem]{Lemma}
\newtheorem{example}[theorem]{Example}

\newcommand{\var}{\mathbb{V}\mathrm{ar}} % Variance
\newcommand{\tr}{\operatorname{tr}}

\newcommand{\bb}[1]{\mathbb{#1}}
\renewcommand{\v}[1]{\mathbold{#1}}
\newcommand{\m}[1]{\mathbf{#1}}

\renewcommand{\c}[1]{\mathcal{#1}}

\begin{document}
\title{Variance Reduction for  Matrix Computations with Applications to Gaussian Processes}
\author{
Anant Mathur$^{1}$ \and Sarat Moka$^{2}$ \and Zdravko Botev$^{1}$\\[0.5em]
\small $^{1}$University of New South Wales, Sydney, NSW 2052, Australia\\
\small $^{2}$Macquarie University, Macquarie Park, NSW 2113, Australia\\
\small \texttt{anant.mathur@unsw.edu.au}
}
\date{}
\maketitle
\begin{abstract}
In addition to recent developments in computing speed and memory, methodological advances have contributed to significant gains in the performance of stochastic simulation. In this paper we focus on variance reduction for matrix computations via matrix factorization. We provide insights into existing variance reduction methods for estimating the entries of large matrices. Popular methods do not exploit the reduction in variance that is possible when the matrix is factorized. We show how computing the square root factorization of the  matrix can achieve in some important cases arbitrarily better stochastic performance. In addition, we detail a factorized  estimator for the trace of a product of matrices and numerically demonstrate that the estimator can be up to 1,000 times more efficient on certain problems of estimating the log-likelihood of a Gaussian process. Additionally, we provide a new estimator of the log-determinant of a symmetric positive-definite matrix where the log-determinant is treated as a normalizing constant of a probability density.
\medskip

\end{abstract}

\begin{center}
\small
This is an author-prepared version of the work published in
\emph{Performance Evaluation Methodologies and Tools (VALUETOOLS 2021)},
LNICST 404, pp.~243--261. The Version of Record is available at
\url{https://doi.org/10.1007/978-3-030-92511-6_16}.
\end{center}
%
%
%In addition to recent developments in computing speed and memory, significant gains in the performance of stochastic simulation are attributed to methodological advances. In this paper, we provide insights into existing variance reduction methods for estimating the entries of large matrices. However, popular methods of estimation do not exploit the reduction in variance that is possible when the matrix is factorized. We show how computing the square root factorization of the  matrix can achieve in some important cases arbitrarily better stochastic performance. In addition, we propose a factorized  estimator for the trace of a product of matrices and numerically demonstrate that the estimator can be up-to 1,000 times more efficient on certain  problems of estimating the log-likelihood of a Gaussian process. Additionally, we provide a new estimator of the log-determinant of a symmetric positive-definite matrix where the log-determinant is treated as a normalizing constant of a probability density.
%
\section{Introduction}
In many scientific computing and machine learning problems we are required to estimate the entries of a positive semi-definite (PSD) matrix $\m A=(a_{i,j})\in\bb{R}^{n\times n}$, where $\m A$ is not explicitly available, but is rather only available as a black box function. That is to say, we assume we only have access to an oracle function that for any $\v x\in\bb{R}^{n}$ can compute $\m A\v x$. Current methods \cite{bekas2007estimator}  employ Monte Carlo simulations to produce unbiased estimates for $a_{i,j}$ via operations of the kind $\m A\v Z$ where $\v Z\in\bb{R}^{n}$ is a random vector drawn from a pre-defined distribution. Hutchinson \cite{hutchinson1989stochastic} was the first to use this methodology to approximate matrix traces, $\tr(\m A) = \sum_i a_{i,i}$ and Bekas et al. \cite{bekas2007estimator} extended  this method to estimate the actual diagonal values $a_{i,i}$. Significant innovations \cite{avron2011randomized, meyer2021hutch++, fitzsimons2016improved, adams2018estimating, stathopoulos2013hierarchical} in this area of research have sought to improve the performance of these estimators through variance reduction and other means, effectively improving their scalability to larger matrices.

One major application of this kind of estimation is fitting Gaussian processes where one is required to evaluate the functions of the black box matrices: $\tr(\m K^{-1}\frac{\partial\m K}{\partial\theta})$ and $\ln|\m K|=\tr(\ln(\m K))$, where $\m K\in\bb{R}^{n\times n}$ is a covariance matrix parameterized by $\theta$, and is of dimension $n$ - the size of the data-set. In these cases, explicitly computing the matrix functions roughly requires $O(n^3)$ operations, making their evaluation infeasible for any matrix of reasonable size, whereas, $\m K^{-1}\v x$ and $\log(\m K)\v x$ can be estimated in $m$ iterations via a Krylov Subspace method that takes $O(mn^2)$ time. Generally, $m\ll n$ iterations are required with each iteration only requiring a matrix-vector-multiplication operation with $\m K$. The scope of black box matrix estimation, however, is not limited to Gaussian processes. For example, one can estimate statistical leverages for a sparse linear model by approximating the diagonals of the projection matrix $\m A = \m X(\m X^{\top}\m X)^{-1}\m X^{\top}$ as in \cite{drineas2012fast} or one can precondition neural networks by estimating the diagonals of a Hessian matrix \cite{dauphin2015equilibrated}. Furthermore, black box trace estimation has many applications including but not limited to evaluating the Cramer-Rao lower bound for the variance of an unbiased estimator \cite{casella2021statistical} and approximating spectral densities \cite{lin2016approximating}.

Martens et al. \cite{martens2012estimating} suggest that the variance for the Monte Carlo estimate of $a_{i,j}$ can depend on the matrix factorization of $\m A$, and our work extends their ideas further. Our paper explores how taking the square root factorization of $\m A=\m A^{1/2}\m A^{1/2}$ can reduce variance with little to no additional cost to CPU time.

In particular, our paper makes the following contributions:
\begin{enumerate}
    \item We illustrate the advantages of black box matrix estimation with factorization over estimation without factorization. In particular, we consider the estimator of Bekas et al. \cite{bekas2007estimator} for the diagonal elements that uses no factorization and compare it with the factorized square root estimator to show that the former can exhibit an arbitrarily larger variance compared to the latter. No such analysis has been given in \cite{martens2012estimating}. We further provide analysis on how the distribution of $\v Z$ effects the variance of the factorized estimator. This includes  variance results for when $\v Z$ are simulated to be uniformly distributed on the surface of the unit radius $n$-dimensional sphere centered at the origin. 
    \item We introduce an estimator for the trace of a product of matrices, $\tr\left(\m A \m W\right)$ where $\m A\in\bb{R}^{n\times n}$ is a PSD matrix and $\m W\in\bb{R}^{n\times n}$ is a symmetric matrix, see also \cite{stein2013stochastic} and \cite{geoga2020scalable}. This estimator exploits square root factorization to achieve variance reduction. We provide numerical examples for the problem of estimating the log-likelihood of a Gaussian process to suggest that the estimator can be up to 1,000 times more efficient in performance, compared to existing methods, for commonly used covariance matrices with dimension $n$  up to 20,000.
    \item We propose a novel Monte Carlo estimator for the log-determinant, $\ln |\m A|$ where $\m A\in\bb{R}^{n\times n}$ is assumed to be a symmetric positive-definite matrix. Existing Monte Carlo methods for approximating the log-determinant require evaluating expressions such as $\ln(\m K)\v Z$ via the Lanczos algorithm \cite{golub2013matrix,ubaru2017fast}. Our proposed method instead relies on the more memory efficient operation of solving the linear-system $\m K^{-1} \v Z$ via the conjugate gradient method.
\end{enumerate}
This paper is organized as follows. In Section 2 we review how one can estimate the entries of the PSD matrix $\m A$ via matrix vector multiplications with the random vector $\v Z$. We then explain how factorization can attain variance reduction and we provide three examples to illustrate when factorization provides greater performance. In this section, we also discuss how one can efficiently compute the factorized estimator. In Section 3, we introduce a Monte Carlo estimator for $\tr(\m A\m W)$ and we illustrate the performance improvement attained when the estimator is applied to the problem of estimating the log-likelihood of a Gaussian process. In Section 4 we discuss the existing Monte Carlo method for estimating $\ln|\m A|$ via the Lanczos algorithm \cite{ubaru2017fast} and we introduce a new Monte Carlo estimator which treats the log-determinant as the normalizing constant of a probability density. Finally, we discuss implications of our research and areas of future work. All proofs are available in the Appendix unless stated otherwise.

\section{Estimating Matrix Elements}
Suppose we wish to estimate the entries of the PSD matrix $\m{A}=(a_{i,j})\in\bb R^{n\times n}$. Let $\m B\m C^\top$ be an arbitrary matrix factorization of the matrix $\m{A}$, with $\m B,\m C \in\bb{R}^{n\times p}$. Let $\v Z\in\bb{R}^{p}$ be a random vector such that $\bb{E}\left[\v Z \v Z^{\top}\right]=\m I_{p}$. Then, a rank-1 unbiased estimate of $\m A$ is given in \cite{martens2012estimating}:
\begin{equation}
\label{bekas estimator}
\hat{\m A} = \m B\v Z \v Z^{\top}\m C^{\top} = \left(\m B \v Z\right)\left(\m C \v Z\right)^{\top}.
\end{equation}
To approximate $a_{i,j}$ we evaluate $\left[\m B \v Z\right]_i\times\left[\m C \v Z\right]_j$, and to approximate the diagonal of $\m A$ we evaluate $\left(\m B \v Z\right)\odot\left(\m C \v Z\right)$, where $\odot$ is defined as the element-wise matrix product. If we select the trivial factorization, $\m A=\m A \m I_n$, that is when $\m B= \m A$ and $\m C = \m I_n$, we obtain the well known Bekas et al. \cite{bekas2007estimator} estimator $\hat{\ell}_{\text{Bekas}}$ for the diagonal,
\begin{equation}
\label{bekas diagonal estimator}
   \hat{\ell}_{\text{Bekas}}= \v Z\odot\m A\v Z.
\end{equation}
Various sampling distributions for $\v Z$ have been proposed, including the Rademacher distribution, whose entries are either $-1$ or $1$ each with probability $0.5$. One can also sample $\v Z$ from the standard Gaussian, uniformly from the set of standard basis vectors or uniformly from the surface of the unit radius $n$-dimensional sphere centered at the origin. 

We note that Bekas et al. \cite{bekas2007estimator} suggest scaling the diagonal estimate $\hat{\ell}_{\text{Bekas}}$ element-wise by the vector $\v Z\odot\v Z$. This scaling has no effect when $\v Z$ is Rademacher as $\v Z\odot\v Z$ is equal to the vector of ones. When $\v Z$ is Gaussian, the performance with scaling is improved, as proven by Kaperick \cite{kaperick2019diagonal}. However, even with scaling the variance of the Gaussian estimator is still marginally higher than that of the Rademacher estimator.

The estimator $\hat{\ell}_{\text{Bekas}}$ does not guarantee non-negative diagonal entries when $\m A$ is a PSD matrix. To mitigate this, we can use a decomposition of the form $\m A=\m B\m B^\top$, that is, when $\m C = \m B$. The entries of the ensuing estimate $\m B \v Z \odot \m B \v Z$ are now guaranteed to be non-negative. We note that this decomposition is not unique and can for example be the Cholesky decomposition, or the symmetric square root factorization $\m A = \m A^{1/2}\m A^{1/2}$. Moreover, if the functional form of $\m A$ is known, it may be exploitable and the factorization $\m A=\m B\m B^{\top}$ can be constructed without computing the symmetric square root or Cholesky factorization.

\begin{example}[Diagonals of Inverse Matrices]\rm
 Suppose we wish to estimate the variances for the OLS estimator $\hat{\v \beta} = (\m X^{\top}\m X)^{-1}\m X^{\top}\v y$, where $\m X\in \bb R^{n\times p}$, and $\v y \in \bb{R}^{n}$. Assume our observations are independent and $\var(y_{i})=\sigma^2$ for all $i=1,\dots,n$. Then, the covariance matrix $\m A = \m \var\left[\hat{\v \beta}\right]$ is, 
\[
\m A = \sigma^2(\m X^{\top}\m X)^{-1}\m X^{\top}\m I_{n} \m X(\m X^{\top}\m X)^{-1}= \sigma^2(\m X^{\top}\m X)^{-1}.
\]
Our goal is to estimate the diagonal entries of $\m A$. In this example, we do not need to compute the square root of $\m A$ directly (e.g., $\m B=\m A^{1/2}$). Instead, we can use conjugate gradient methods, since $\m A=\m B\m B^\top$, where,
\[
\m B=\sigma(\m X^\top\m X)^{-1}\m X^\top.
\] 
Thus, the factorized diagonal estimator is $\hat\ell=\m B \v Z\odot\m B\v Z$ and the computation $(\m X^\top\m X)^{-1}\v c$, where $\v c = \m X^{\top}\v Z$ can be solved via the conjugate gradient method.
\end{example}
\subsection{Variance Analysis}
Martens et al. \cite{martens2012estimating} provides variance formulas for estimator \eqref{bekas estimator} when $\v Z$ is drawn from the standard Gaussian and Rademacher distribution. In this section we state these formulas along with the variance formula for when $\v Z$ is drawn uniformly on the surface of the unit radius $n$-dimensional sphere centered at the origin. For the following analysis we assume that $\hat{\m A} = (\hat{a}_{i,j}) = \left(\m B \v Z\right)\left(\m C \v Z\right)^{\top}$ and $\m B, \m C\in \bb{R}^{n\times n}$. Furthermore, we denote $\v b_{i,:}$ and $\v c_{i,:}$ as the column vectors containing the $i$-th rows of the matrices $\m B$ and $\m C$.
\noindent
\subsubsection{Gaussian:}
When $\v Z\sim\mathcal{N}(\mathbf{0}, \m I_{n})$, the variance for $\hat{a}_{i,j}$ is,
\begin{equation}
\label{bekas_var}
\var_G[\hat{a}_{i,j}]= \|\v b_{i,:}\|^2\|\v c_{j,:}\|^2+a^2_{i,j},    
\end{equation}
and when $\m B=\m A$, $\m C=\m I_{n}$, and $i=j$, we obtain the variance for the diagonal estimator of Bekas et al. \cite{bekas2007estimator}, that is,
\begin{equation}
\var_G^{\m A \m I}[\hat{a}_{i,i}]= 2a^2_{i,i}+\sum_{j\neq i}a^2_{i,j}.
\end{equation}

\begin{theorem}[Theorem 4.1 in Martens et al. \cite{martens2012estimating}]\label{min_gauss}
For the factorization $\m A = \m B \m C^{\top}$, $\var_G[\hat{a}_{i,i}]$ is minimized when $\m C = \m B$ with resulting variance,
\begin{equation}
\var_G^{\m B \m B^{\top}}[\hat{a}_{i,i}]= 2a^2_{i,i}.
\end{equation}
\end{theorem}
Our next result contributes to our understanding of using a spherical probing vector.  
\subsubsection{Uniform Spherical:}
Suppose $\v Y\sim\mathcal{N}(\mathbf{0}, \m I_{n})$ then $\v Z = \v Y/\|\v Y\|$ is uniformly distributed on the surface of the unit radius $n$-dimensional sphere centered at the origin. 
\begin{theorem}[Spherical Estimator for Matrix Elements]\label{sphere_element_var}
Suppose $\v Z$ is uniformly distributed on the surface of the unit radius $n$-dimensional sphere centered at the origin and $\m A= \m B\m C^{\top}$. Then,
\[\bb{E}\left[n\left(\m B \v Z\right)\left(\m C \v Z\right)^{\top}\right] = \m A,\]
and $\hat{a}_{i,j}=n\left[\m B \v Z\right]_i\times\left[\m C \v Z\right]_j$ has variance,
\[\var_S[\hat{a}_{i,j}]=\frac{n}{n+2}\|\v b_{i,:}\|^2\|\v c_{j,:}\|^2+\frac{n-2}{n+2}a_{i,j}^2.\]
\end{theorem}
As we can see, when $n$ is large, $\var_S[\hat{a}_{i,j}]\approx\var_G[\hat{a}_{i,j}]$. 
\subsubsection{Rademacher:}
Martens et al. \cite{martens2012estimating} give the following variance for $\hat{a}_{i,j}$ when $\v Z \in\bb{R}^n$ is Rademacher, 
\begin{align*}
\var_R[\hat{a}_{i,j}]&=\var_G[\hat{a}_{i,j}]-2\sum_k b^2_{i,k}c^2_{j,k}.\\
&= \|\v b_{i,:}\|^2\|\v c_{j,:}\|^2+a^2_{i,j}-2\sum_k b^2_{i,k}c^2_{j,k}.
\end{align*}
For the two cases where 1) $\m B=\m A$, $\m C=\m I$ (unfactorized) and 2) $\m C=\m B$ (factorized), we have the variances for the diagonal entry estimates,
\begin{align}
\var_R^{\m A \m I}[\hat{a}_{i,i}]&= \sum_{j\neq i}a^2_{i,j}.\\
\var_R^{\m B \m B^{\top}}[\hat{a}_{i,i}]&= 2a^2_{i,i}-2\sum_k b^4_{i,k}.
\end{align}
These formulas indicate $\var_R^{\m B \m B^{\top}}\leq\var_G^{\m B \m B^{\top}}$ and $\var_R^{\m A \m I}[\hat{a}_{i,i}]\leq\var_G^{\m A \m I}[\hat{a}_{i,i}]$; therefore we recommend the Rademacher as the best choice for $\v Z$.

\subsubsection{Extending Results on Rademacher:}
Theorem \ref{min_gauss} shows that $\m A = \m B \m B^{\top}$ is the most optimal factorization when $\v Z$ is Gaussian. This is not guaranteed for the Rademacher case and our contribution is to analyze the effect of the square root factorization in the Rademacher case. For example, for PSD matrices with growing off-diagonals with increasing size, $\var_R^{\m A \m I}[\hat{a}_{i,i}]$ can be made arbitrarily large, while $\var_R^{\m B \m B^{\top}}[\hat{a}_{i,i}]$ remains bounded above by $2a_{i,i}^2$. We now show an example matrix that behaves this way.
\begin{example}[Stochastic matrices]
\label{stochastic_matrices}
\normalfont
Suppose we have $x, y$ such that $x > y \geq 0$. Consider the symmetric matrix $\m A$ of size $n\times n$ with
\[
a_{i,j} = \begin{cases}
		x, \quad &\text{ if}\, i = j,\\
		y, \quad &\text{ if} \, i \neq j.
	    \end{cases}
\]
For this matrix $\m A$, we can show that the eigenvalues (or singular values) are $x + py$ with multiplicity $1$ and $x- y$ with multiplicity $n -1$. Since all the eigenvalues are positive, $\m A$ is a positive definite matrix. In particular, 
\[
\var_R^{\m A \m I}[\hat{a}_{i,i}]  - \var_R^{\m B \m B^{\top}}[\hat{a}_{i,i}] \geq \sum_{j \neq i} a^2_{i,j} - 2a^2_{i,i} = (n-1) y^2 - 2 x^2.
\]
Therefore, $\var_R^{\m A \m I}[\hat{a}_{i,i}]  \geq \var_R^{\m B \m B^{\top}}[\hat{a}_{i,i}]$ for all $i = 1, \dots, n$ if $y \geq x\sqrt{2/(n-1)}$. Furthermore, the factorized estimator is substantially better than the unfactorized estimator when the ratio $\sqrt{(n-1)}\,y/x \to \infty$ as $n \to \infty$.\\
For instance, let $\varepsilon_n = \frac{1}{(n-1)^{\gamma}}$ for some constant $1/2 < \gamma < 1$. Then, with $x = \varepsilon_n$ and $y = \frac{1 - \varepsilon_n}{n-1}$, the matrix $\m A$ is a doubly stochastic matrix for all $n$, because each row and each column sum to $x + (n-1)y = 1$. Note that $x > y$ for sufficiently large values of $n$. Furthermore,  
\[
\frac{\sqrt{(n-1)}\, y}{x} = \frac{\sqrt{n-1}}{(n- 1)^{-\gamma}} \frac{1 - \varepsilon_n}{n - 1} = \left(1 - \varepsilon_n\right)(n- 1)^{\gamma -1/2},
\]
which goes to $\infty$ as $n \to \infty$. In other words, the square root estimator can be arbitrarily better than the naive one. 
\end{example}
We now show an example where the unfactorized estimator exhibits an arbitrary larger variance compared to the factorized estimator when estimating an arbitrary matrix element.
\begin{example}[Toeplitz matrices with exponential decay]
\label{toeplitz_ex}
Consider the Gram matrix
\[\m{A}=\{k({x}_i,{x}_j)\}_{i,j=1,\dots,n},\]
where $k$ is the Squared Exponential kernel parameterized by $\v \theta = ( \theta_1, \theta_2)^{\top}$,
\[k({x}_i,{x}_j)=\theta_1\exp\left[\frac{-({x}_i-{x}_j)^2}{2\theta_2^2}\right],\]
for $x_1,\dots,x_n\in\mathbb{R}$. For simplicity let $\theta_1=1$, forcing the diagonal entries to be unitary. Suppose $\{{x}_i\}_{i=1,\dots,n}$ are equally separated with spacing $\frac{1}{n-1}$, then $\m{A}$ is Toeplitz. The entries of $\m{A}$ can be written as,
\[a_{i,j}=\exp\left[\frac{-(i-j)^2}{2(n-1)^2\theta_2^2}\right].\]
For the estimate $\hat{a}_{i,j}$, the variance of the unfactorized estimator has the following lower bound, 
\begin{align*}
\var_R^{\m A \m I}[\hat{a}_{i,j}]&= \sum_{k=1}^{n}\exp\left[\frac{-(i-k)^2}{(n-1)^2\theta_2^2}\right]-\exp\left[\frac{-(i-j)^2}{(n-1)^2\theta_2^2}\right]\\
&\geq n\exp(-\theta_2^{-2})-1.
\end{align*}
If we let $\m B = (b_{i,j})\in\mathbb{R}^{n \times n}$ be the square root matrix of $\m A$, then we obtain the upper bound for the variance of the factorized estimator,
\begin{align*}
\var_R^{\m B \m B^{\top}}[\hat{a}_{i,j}]&= 1+a_{i,j}^2-2\sum_m b^2_{i,m}b^2_{j,m}\leq 1+a_{i,j}^2\leq2.
\end{align*}
In this example, $\var_R^{\m A \m I}[\hat{a}_{i,j}]$ can be made arbitrary larger than the variance for the factorized estimator. That is to say, for all $\delta>0$ there exists $n'$ such that $\var_R^{\m A\m I}[\hat{a}_{i,j}]-\var_R^{\m B \m B^{\top}}[\hat{a}_{i,j}]>\delta$ for all $n>n'=\lceil(\delta +3)\exp(\theta_2^{-2})\rceil$. 
\end{example}
We now show an example of the matrix $\m A$ where the unfactorized Rademacher estimator exhibits lower variance than the factorized estimator.
\begin{example}[Tridiagonal matrices] It is clear that a PSD tridiagonal matrix should be an example where we expect the unfactorized estimator to perform on par or better than the symmetrized estimator due to its minimal off-diagonal entries. While it is safe to assume that this case is rare in application, it is still a worthwhile example to show when factorizing $\m A$ may not be advantageous. Let $\m A\in\bb{R}^{n\times n}$ be the symmetric tridiagonal matrix, 
\[\m{A} = \begin{bmatrix}1&c\\c&1&c\\&c&\ddots &\ddots \\&&\ddots &\ddots &c\\&&&c&1\end{bmatrix}.\]
Then, the Cholesky decomposition of $\m A$ is $\m{A}=\m B\m B^{\top}$, where $\m B\in\bb{R}^{n\times n}$ is a lower bi-diagonal matrix with entries given by the following recursive formulas,
\begin{equation}
\label{chol_eqn}
b_{i,i}=\sqrt{1-b_{i,i-1}^2},\quad \text{and}\quad b_{i,i-1}=\frac{c}{b_{i-1,i-1}},
\end{equation}
where $b_{1,1}=1$ and $b_{2,1}=c.$ By using (6), (7) and \eqref{chol_eqn} we can show,
\begin{align}
\label{chol_limit}
\var_R^{\m B \m B^{\top}}[\hat{a}_{i,i}] &= 4c^2\left[\frac{1}{b_{i-1,i-1}^2}-\frac{c^2}{b_{i-1,i-1}^4}\right] = 4c^2 \frac{b_{i,i}^2}{b_{i-1, i-1}^2},\quad \text{and}
\end{align}
\[
\var_R^{\m A \m I^{\top}}[\hat{a}_{i,i}] = \begin{cases}
		c^2, \quad &\text{ if}\,i = 1\text{ or }n,\\
		2c^2, \quad &\,\text{otherwise}.
	    \end{cases}
\]
By numerically evaluating \eqref{chol_eqn} we observe that $\frac{b_{i,i}^2}{b_{i-1, i-1}^2}$ is non-decreasing in $i$, hence $\var_R^{\m B \m B^{\top}}[\hat{a}_{i,i}]$ is non-decreasing in $i$. We further observe that as the limit $n\rightarrow\infty$, 
\[\var_R^{\m B \m B^{\top}}[\hat{a}_{n,n}]\uparrow 4c^2.\]
Based on these numerical observations, 
\[
\frac{\var_R^{\m B \m B^{\top}}[\hat{a}_{i,i}]}{\var_R^{\m A \m I}[\hat{a}_{i,i}]} \leq \frac{\var_R^{\m B \m B^{\top}}[\hat{a}_{n,n}]}{\var_R^{\m A \m I}[\hat{a}_{n,n}]},
\]
and as 
$n\rightarrow\infty$,
\[\frac{\var_R^{\m B \m B^{\top}}[\hat{a}_{n,n}]}{\var_R^{\m A \m I}[\hat{a}_{n,n}]}\uparrow 4.\]
\end{example}
Therefore, $\var_R^{\m B \m B^{\top}}[\hat{a}_{i,i}]$ will be at most $4\times\var_R^{\m A \m I}[\hat{a}_{i,i}]$ for large values of $n$. 

Numerical simulations with the square root factorization $\m B = \m A^{1/2}$ indicate a similar conclusion. In particular, for certain values of $c$, $\var_R^{\m A \m I}[\hat{a}_{i,i}]>\var_R^{\m B \m B^{\top}}[\hat{a}_{i,i}]$, with a relative difference of at most 0.4. This relative difference remains bounded above by 0.4 as $n\rightarrow\infty$, whereas in Examples \ref{stochastic_matrices} and \ref{toeplitz_ex} the relative difference, when the factorized estimator outperforms the unfactorized estimator, goes to $\infty$ as $n\rightarrow\infty$. In plain English, while the downside of using the factorized estimator is limited, the upside can be unlimited as the size of the matrix grows.

\subsection{Computing $\m A^{1/2} \v Z$}
\label{Lanczos methods}
We briefly review well-known methods for  efficiently computing with the square root factorization. As discussed in the introduction, black box estimation relies on the ability to efficiently evaluate matrix-vector operations. Thus, when using the factorization $\m A =\m B \m B^{\top}$ we must consider how to evaluate $\m B \v Z$. If the functional form of $\m A$ is not exploitable or not known, the only practical choice for $\m B$ is $\m B = \m A^{1/2}$, the unique symmetric square root of $\m A$.

Chow and Saad \cite{chow2014preconditioned} outline how to compute $\m A^{1/2} \v Z$ via the well-known Lanczos tridiagonalization algorithm. The Lanczos algorithm takes the black box matrix $\m A$ and the initial vector $\v Z$ and after $m\leq n$ iterations outputs the matrices $\m T_m$ and $\m V_m$ that satisfy, 
\[\m T_m = \m V_m^{\top} \m A \m V_{m},\]
where $\m V_m\in\bb{R}^{n\times m}$ contains orthonormal columns, and $\m T_m \in \bb{R}^{m\times m}$ is a symmetric tridiagonal matrix.  We can then use the approximation $\m V_m^{\top} f(\m A )\m V_{m}\approx f\left(\m V_m^{\top} \m A \m V_{m}\right)=f(\m T_m)$ and take $f(\m A) = \m A^{1/2}$ to obtain the approximation, 
\begin{equation}
    \v x_m^*=\|\v Z\|\m V_{m}\m T^{1/2}_{m}\v e_1.
\end{equation}
Generally $\|\v x_m^*-\m A^{1/2}\v Z\|$ can be made small when $m \ll n$, thus allowing efficient computation and storing of $\m T^{1/2}_{m}$. A discussion on the convergence rates for the Lanczos approximation can be found in Chow and Saad \cite{chow2014preconditioned}. In general, this method has $O(mn^2)$ time complexity.

Alternatively, we can evaluate $\m A^{1/2} \v Z$  using the contour integral quadrature (CIQ) of Pleiss et al. \cite{pleiss2020fast}. This method also offers $O(mn^2)$ cost where $m$, the number of iterations is often less than 100 even for matrices of size up to 50,000. This method can be implemented via the \href{https://gpytorch.ai/}{GPyTorch} framework \cite{gardner2018gpytorch}.

\section{Estimating the Trace of a Product of Matrices}
In this section we introduce an estimator for the trace of a product of matrices, $\tr\left(\m A \m W\right)$, where $\m A\in\bb{R}^{n\times n}$ is a PSD matrix and $\m W\in\bb{R}^{n\times n}$ is a symmetric matrix. Applications of such traces abound and an example with Gaussian processes is  given later on. If the goal is to estimate $\tr(\m A)$, we can take the sum of the estimator $\hat{\ell}_{\text{Bekas}}$ to obtain the well known Hutchinson estimator,
\begin{equation}
{\tr(\hat{\m A})} = \v Z^{\top}\m A\v Z,
\end{equation}
where $\bb{E}[\v Z\v Z^{\top}] = \m I_n$.
For the symmetric square root $\m B=\m A^{1/2}$, summing the factorized
diagonal estimates gives exactly the Hutchinson trace estimate:
\[
\sum_i [\m B\v Z]_i^2
=\|\m B\v Z\|^2
=\v Z^\top\m B^\top\m B\v Z
=\v Z^\top\m A\v Z.
\]
Here $\m B^\top\m B=\m A$ because $\m B$ is the symmetric square root.
For a general factorization $\m A=\m B\m B^\top$, the sum is
$\v Z^\top\m B^\top\m B\v Z$. Its expectation is still $\tr(\m A)$
when $\bb E[\v Z\v Z^\top]=\m I$, but it need not equal
$\v Z^\top\m A\v Z$ for each draw.
\par
However, suppose we want to estimate $\tr(\m A \m W)$, where $\m A\in\bb{R}^{n\times n}$ is a PSD matrix and $\m W \in \bb{R}^{n \times n}$ is a symmetric matrix (not necessarily PSD). We can then use the following estimator,
\begin{equation}
\label{sym_trace}
\hat{\psi}_{\mathrm{sqrt}} = \v Z^\top \m B^\top \m W\m B\v Z,
\end{equation}
where $\m B = \m A^{1/2}$, see also \cite{stein2013stochastic} and \cite{geoga2020scalable}. To see this is unbiased, observe that due to the invariant cyclic nature of the trace,
\[\bb{E}\left[\v Z^\top \m B^\top \m W\m B\v Z\right]=\bb{E}\left[\tr(\m Z\m Z^{\top}\m B^\top\m W\m B)\right] = \tr(\m A \m W).\]
\subsection{Variance Analysis}
We now provide comparative variance analysis of the square root estimator \eqref{sym_trace} when $\v Z$ is drawn from the standard Gaussian, from the Rademacher distribution, and uniformly on the surface of the unit radius $n$-dimensional sphere centered at the origin. For the following analysis we refer to the square root estimator as,
\[\hat{\psi}_{\mathrm{sqrt}} = \v Z^\top \m B^\top \m W\m B\v Z,\]
where $\m B = \m A^{1/2}$, and the standard Bekas et al. \cite{bekas2007estimator} estimator as,
\[\hat{\psi} = \v Z^\top \m A\m W\v Z.\]

\subsubsection{Gaussian}
\begin{theorem}[Lemma 9 in Avron and Toledo \cite{avron2011randomized}]
\label{gauss_quad_var}
Suppose $\m \Sigma\in\bb{R}^{n\times n}$, and $\v Z\sim\mathcal{N}(\mathbf{0}, \m I_{n})$. Let $\m \Sigma' = (\m\Sigma + \m\Sigma^{\top})/2$. Then,
\begin{equation}
    \var\left[{\v Z^{\top}\m\Sigma\v Z}\right]= 2\|\m\Sigma'\|_{\operatorname{F}}^2,
\end{equation}
where $\|\m X\|_{\operatorname{F}}:=\sqrt{\tr(\m X^{\top}\m X)}=\sqrt{\sum_{i,j}\m (\m X_{i,j})^2}$.
\end{theorem}
Following from this theorem we have the following.
\begin{lemma}[Comparing the Variance of the Gaussian Trace Estimator]
If $\v Z\sim\mathcal{N}(\mathbf{0}, \m I_{n})$, then $\bb{E}[\hat{\psi}_{\mathrm{sqrt}}]=\bb{E}[\hat{\psi}]=\tr(\m A\m W)$,
\begin{align}
\label{gauss_f1}
&\var_G[\hat{\psi}]= \tr\left((\m A\m W)^\top(\m A\m W)\right)+\tr\left((\m W\m A)^\top(\m A\m W)\right), \quad \text{and}\\
\label{gauss_f2}
&\var_G[\hat{\psi}_{\mathrm{sqrt}}] = 2\tr\left((\m W\m A)^\top(\m A\m W)\right).
\end{align}
Furthermore,
\[\var_G[\hat{\psi}_{\mathrm{sqrt}}]\leq\var_G[\hat{\psi}].\]
\end{lemma}
\begin{proof}
Using Theorem \ref{gauss_quad_var}, we obtain the variance formulas \eqref{gauss_f1} and \eqref{gauss_f2}. To prove $\var[\ell_{\mathrm{sqrt}}]\leq\var[\hat{\psi}]$ we need to show,
\[
\tr\left((\m W\m A)^\top(\m A\m W)\right)\leq\tr\left((\m A\m W)^\top(\m A\m W)\right).
\]
As $\left\langle\m X,\m Y\right\rangle_{\operatorname{F}} := \tr(\m X^\top \m Y)$ defines the Frobenius inner-product, the inequality holds true from the Cauchy–Schwarz inequality.
\end{proof}
\subsubsection{Uniform Spherical}
\begin{theorem}[Variance of the Spherical Quadratic Form]
\label{sphere_quad_var}
Suppose $\m \Sigma\in\bb{R}^{n\times n}$ and $\v Z$ is uniformly distributed on the surface of the unit radius $n$-dimensional sphere centered at the origin. Let $\m \Sigma' = (\m\Sigma + \m\Sigma^{\top})/2$. Then,
\begin{equation}
\var\left[n\v Z^{\top}\m \Sigma' \v Z\right]=\frac{n}{n+2}\times2\left(\|\m\Sigma'\|_{\operatorname{F}}^2-\frac{(\sum_i\m\Sigma'_{i,i})^2}{n}\right).
\end{equation}
\end{theorem}
Following from this theorem we have the following.
\begin{lemma}[Variance of the Spherical Trace Estimator]
If $\v Z$ is uniformly distributed on the surface of the unit radius $n$-dimensional sphere centered at the origin, then $\bb{E}[n\hat{\psi}_{\mathrm{sqrt}}]=\bb{E}[n\hat{\psi}]=\tr(\m A\m W)$, 
\begin{align}
&\var_{S}\left[n\hat{\psi}\right] = \frac{n}{n+2}\var_{G}\left[\hat{\psi}\right] - \frac{2}{n+2}\left(\sum_i \left[ \frac{\m A\m W+\m W\m A}{2} \right]_{i,i}\right)^2, \quad \text{and}\\
&\var_{S}\left[n\hat{\psi}_{\mathrm{sqrt}}\right]= \frac{n}{n+2}\var_{G}\left[\hat{\psi}_{\mathrm{sqrt}}\right] - \frac{2}{n+2}\left(\sum_{i} [\m B^\top\m W\m B]_{i,i}\right)^2.
\end{align}
\end{lemma}
As we can see,  $\var_S\approx\var_G$ for large values of $n$.
\subsubsection{Rademacher}
\begin{theorem}[Lemma 1 in Avron and Toledo \cite{avron2011randomized}]
\label{rad_quad_var}
Suppose $\m \Sigma\in\bb{R}^{n\times n}$, and $\v Z\in\bb{R}^n$ follows a Rademacher distribution. Let $\m \Sigma' = (\m\Sigma + \m\Sigma^{\top})/2$. Then,
\begin{equation}
    \var\left[{\v Z^{\top}\m\Sigma\v Z}\right]=2\left[\|\m\Sigma'\|_{\operatorname{F}}^2-\sum_{i}{\m\Sigma'}_{i,i}^2\right].
\end{equation}
\end{theorem}
Following from this theorem we have the following.
\begin{lemma}[Variance of the Rademacher Trace Estimator]
If $\v Z$ is a n-dimensional Rademacher random vector, then $\bb{E}[\hat{\psi}_{\mathrm{sqrt}}]=\bb{E}[\hat{\psi}]=\tr(\m A\m W)$,
\begin{align}
&\var_{R}\left[\hat{\psi}\right] = \var_{G}\left[\hat{\psi}\right] - 2 \sum_i \left[ \frac{\m A\m W+\m W\m A}{2} \right]^2_{i,i}, \quad \text{and}\\
&\var_{R}\left[\hat{\psi}_{\mathrm{sqrt}}\right]= \var_{G}\left[\hat{\psi}_{\mathrm{sqrt}}\right] - 2 \sum_{i} [\m B^\top\m W\m B]_{i,i}^2.
\end{align}
\end{lemma}
Analogous to our conclusions for estimating arbitrary matrix elements, the above results indicate that the Rademacher estimator is guaranteed to be better than Gaussian estimator for estimating the trace of $\m A\m W$. Moreover, when $\v Z$ is Gaussian, we have proven $\hat{\psi}_{\mathrm{sqrt}}$  always performs better than $\hat{\psi}$. However, this is not always true when $\v Z$ is Rademacher.

Let $\m \Lambda= (\m A\m W+\m W\m A)/2$, and $\m \Omega = \m B^\top\m W\m B$. Then $\var_{R}\left[\hat{\psi}_{\mathrm{sqrt}}\right]>\var_{R}\left[\hat{\psi}\right]$ if and only if, 
\[\sum_i\left[({\m \Lambda}_{i,i})^2-({\m \Omega}_{i,i})^2\right]>\sum_{i,j}\left[({\m \Lambda}_{i,j})^2-\left({\m \Omega}_{i,j}\right)^2\right].\]
Numerical experiments indicate this inequality can hold when the contribution of $\|\m \Lambda\|_{\operatorname{F}}$ and $\|\m \Omega\|_{\operatorname{F}}$ is concentrated on the diagonal. In these cases both $\hat{\psi}_{\mathrm{sqrt}}$ and $\hat{\psi}$ are very accurate when $\v Z$ is Rademacher. However, when most of the contribution of the Frobenius norm stems from the off-diagonals we observe cases where $\|\m \Lambda\|_2^2\gg\|\m \Omega\|_2^2$ and $\var_{R}\left[\hat{\psi}\right]\gg\var_{R}\left[\hat{\psi}_{\mathrm{sqrt}}\right]$. We now show such an example in the context of maximum likelihood estimation for Gaussian processes.

\subsection{Gaussian Processes}
Gaussian processes (GP's) are used in a variety of modelling applications mainly through the following regression setting. Let $\v X = \{\v x_1,\dots,\v x_n\}$  be a set of $n$ data points and let $\v y = (y_1,\dots,y_n)^{\top}$ be the vector of regression values under the model,
\[y_i = f(\v x_{i})+\epsilon_i,\quad\text{where }\epsilon_i\sim\mathcal{N}(0,\sigma^2).\]
In GP regression, we assume the following prior on the function values $\v f  = (f(\v x_1),\dots,f(\v x_n))^{\top}$,
\[p(\v f|\v{x_1},\dots,\v{x_n})=\mathcal{N}(\mathbf{0},\m K).\]
The covariance matrix $\m K$ has entries $\m{K}_{i,j}=\{k({\v x}_i,{\v x}_j| \v\theta)\}$, where $k$ is a covariance function with hyper-parameters $\v \theta$. A popular choice for $k$ is the Squared Exponential covariance function,
\[k(\v {x}_i,\v {x}_j)=\theta_1\exp\left[\frac{-\|\v x_j-\v x_i\|^2}{2\theta_2^2}\right].\]
To fit a GP we must select the hyper-parameters $\v \theta = (\theta_1,\theta_2)^{\top}$ by maximizing the likelihood. Therefore, we need to evaluate the log-likelihood function,
\begin{equation}
\label{gp-ll}
\ln[\mathcal{L}(\v{\theta}|\v y)]=-\frac{1}{2}\ln|\m K_y|-\frac{1}{2}\v y^{\top}\m K^{-1}_y\v y+\text{constant},
\end{equation}
and the score function,
\begin{equation}
\frac{\partial\ln[\mathcal{L}(\v{\theta}|\v y)]}{\partial{\theta_i}}=-\frac{1}{2}\operatorname{Tr}\left(\m K_y^{-1}\frac{\partial\m K_y}{\partial\theta_i}\right)+\frac{1}{2}\v y^{\top}\m K^{-1}_y\frac{\partial\m K}{\partial\theta_i}\m K^{-1}_y\v y,
\end{equation}
where $\m K_y = \m K + \sigma^2 \m I_{n}$. For large $n$ explicitly evaluating $\tr\left(\m K_y^{-1}\frac{\partial\m K_y}{\partial\theta_i}\right)$ is infeasible. Instead, we must approximate the trace via simulation. Suppose we simulate $M$ draws of the $n$-dimensional Rademacher random variable, $\{\v Z_i\}_{i=1,\dots,M}$. Then, the Hutchinson estimator for $\tr\left(\m K_y^{-1}\frac{\partial\m K_y}{\partial\theta_2}\right)$ is, 
\[
\hat{\psi} = \frac{1}{M}\sum_{i=1}^{M}\left(\m K_y^{-1}\v Z_{i}\right)^{\top}\left(\frac{\partial\m K_y}{\partial\theta_i}\v Z_i\right).
\]
Instead, we can use the square root estimator, 
\[
\hat{\psi}_{\mathrm{sqrt}} = \frac{1}{M}\sum_{i=1}^{M}\v c_i^{\top}\frac{\partial\m K_y}{\partial\theta_i}\v c_i,\quad \text{where }\v c_i=\m K_y^{-1/2}\v Z_i.
\]
\begin{figure}[ht]
  \captionsetup[subfigure]{labelformat=empty}
  \centering
  \subfloat[]{\includegraphics[width=0.49\textwidth]{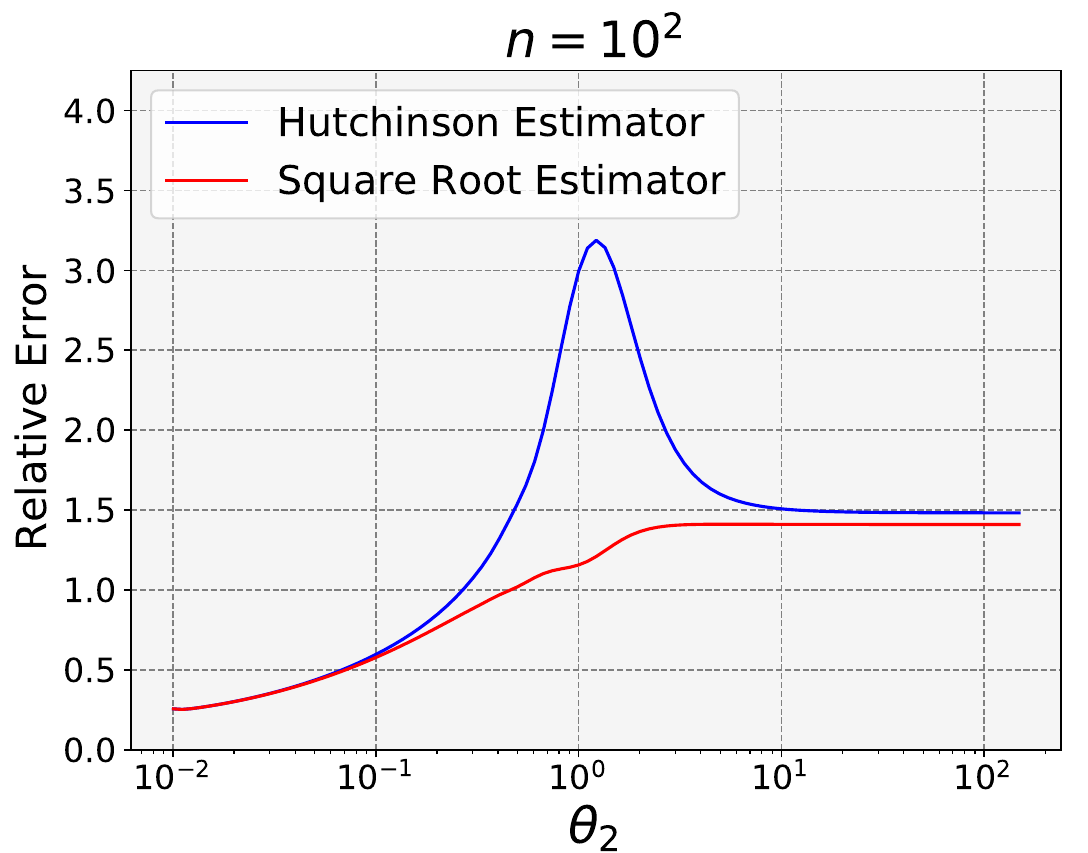}\label{fig:f1}}
  \hfill
  \subfloat[]{\includegraphics[width=0.49\textwidth]{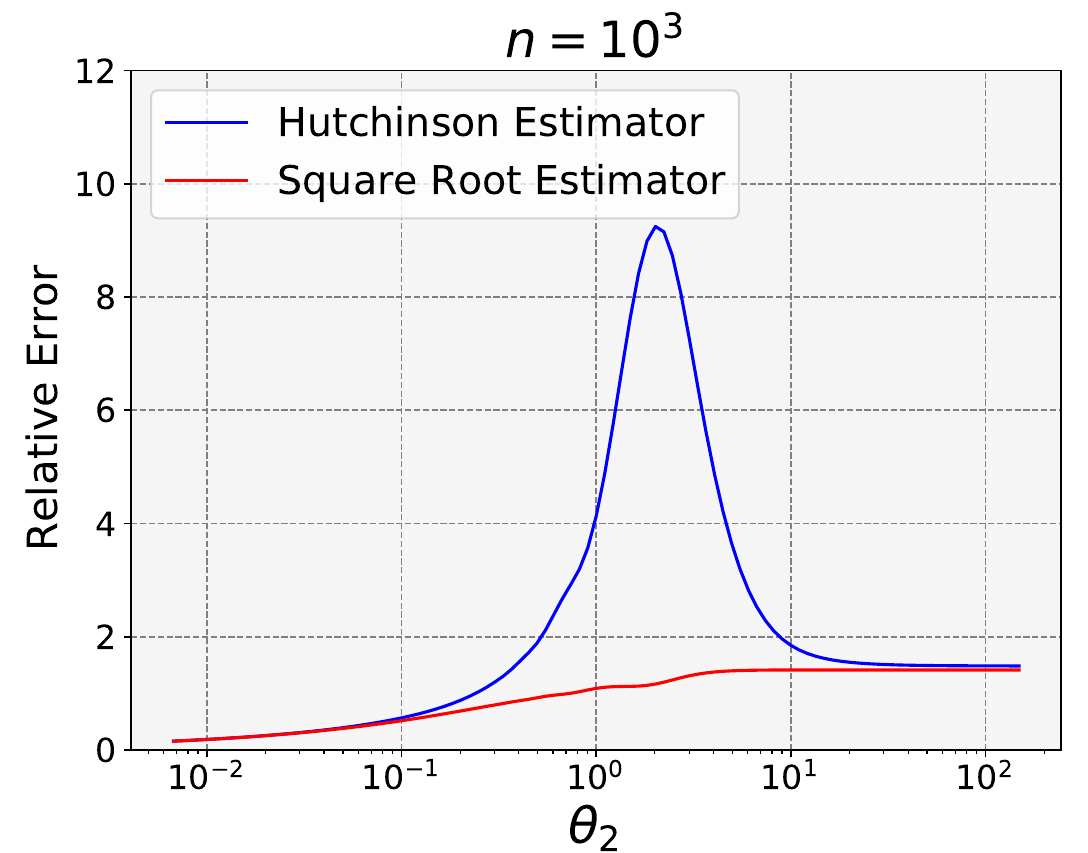}\label{fig:f2}}
  \caption{True relative error as a function of $\theta_2$ for $M=1$ (a single replication), when $n=10^2$ and $n=10^3$. Relative Error $=\sqrt{\var{(\hat{\psi}})}/{|\psi|}$, where $\psi=\tr\left(\m K_y^{-1}\frac{\partial\m K_y}{\partial\theta_2}\right)$.}
\end{figure}
\begin{figure}[ht]
  \captionsetup[subfigure]{labelformat=empty}
  \centering
  \subfloat[]{\includegraphics[width=0.48\textwidth]{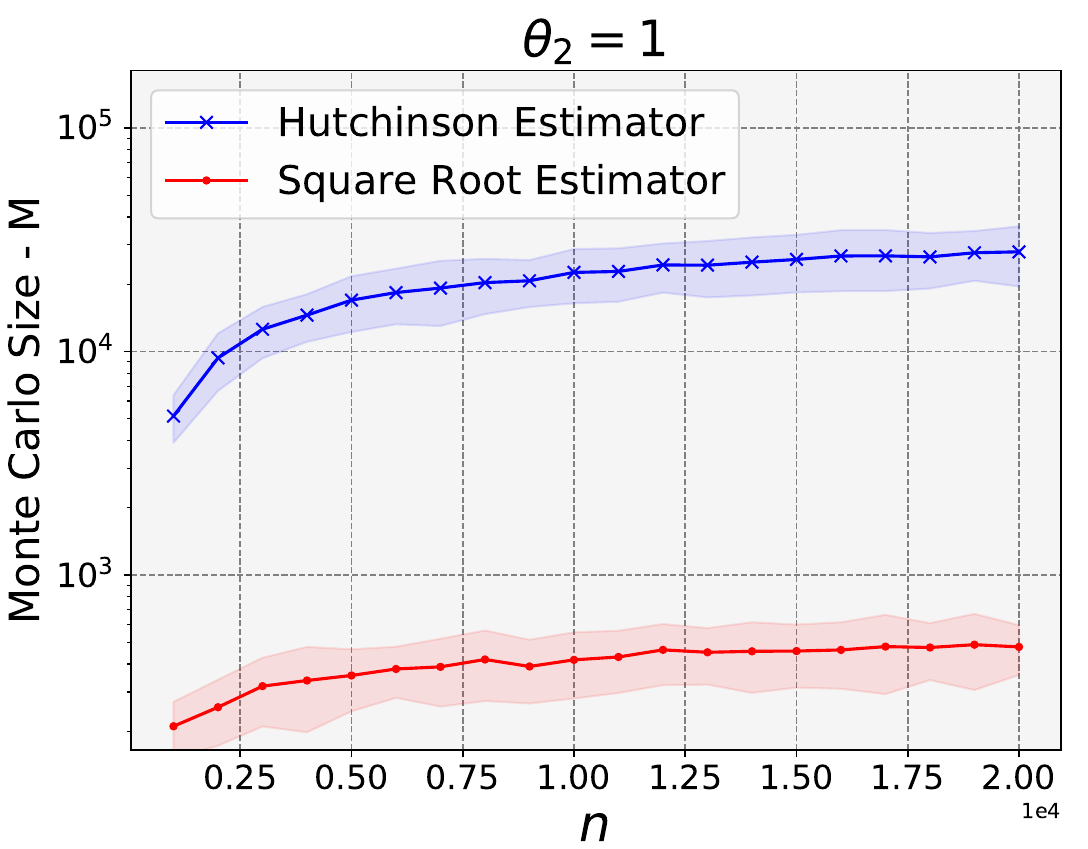}\label{fig:f3}}
  \hfill
  \subfloat[]{\includegraphics[width=0.48\textwidth]{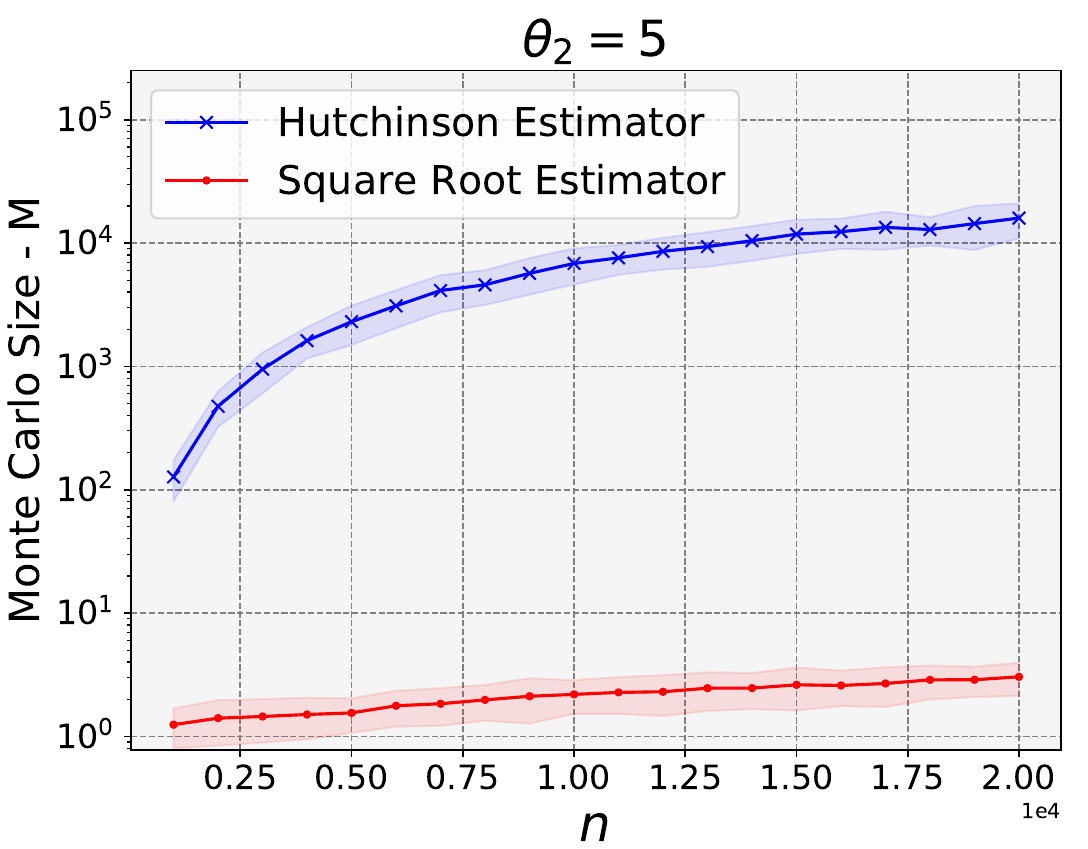}\label{fig:f4}}
  \caption{Mean Monte Carlo size $M$ required in simulation to obtain a mean squared error of $0.25$ as $n$ increases, when $\theta_2=1$ and $\theta_2=5$.}
\end{figure}
We can evaluate $\m K_y^{-1/2}\v Z_i$ with the Pleiss et al. \cite{pleiss2020fast} CIQ method. We observe that on average $\m K_y^{-1/2}\v Z_i$ takes double the CPU time to evaluate as compared to $\m K_y^{-1}\v Z_i$ via the conjugate gradient method; the latter being the required evaluation for the Hutchinson estimator. Importantly, both operations have $O(mn^2)$ complexity, where $m\ll n$.

To compare the performance of the estimators numerically, we fix $\sigma^2 = 0.1$ and $\theta_1 = 1$ and we let $\v X$ be the $n$ equidistant points spanning the interval $[0,1]$. In Fig.1 we evaluate the theoretical relative error of the two estimators when $M=1$ and observe that the square root estimator always performs better than or equal to the Hutchinson estimator. For $\theta_2 \in [10^{-1}, 10^1]$ the relative error for Hutchinson estimator significantly deviates from the relative error of the square root estimator. In Fig.2 we illustrate the differing performance as the size of the matrix increases in size. We observe that when $\theta_2=5$ the Hutchinson estimator can require up-to $1,000$ times more Monte Carlo replications to obtain the same accuracy (mean squared error) than the square root estimator.

\section{Estimating the Log-Determinant}
In this section we propose a conditional Monte Carlo estimator for the log-determinant of a symmetric positive-definite matrix by treating the log-determinant as a normalizing constant of a probability density. As seen from \eqref{gp-ll}, training GP's require the evaluation of the log-determinant of $\m K_y$. Suppose $\m A\in\bb{R}^{n\times n}$ is a symmetric positive-definite matrix. Then, $\ln|\m A|$ can be evaluated in $O(n^3)$ time with the Cholesky decomposition. However, as previously stated this is infeasible for matrices with dimension greater than a few thousand. Existing log-determinant estimation techniques \cite{ubaru2017fast,gardner2018gpytorch}, utilize the Hutchinson estimator via the following approximation,
\begin{equation}
\ln|\m A|=\tr\left(\ln(\m A)\right)=\bb{E}[\v Z^{\top}\ln (\m A)\v Z]\approx\frac{1}{M}\sum_{i=1}^M\v Z_{i}^{\top}\ln (\m A)\v Z_i,
\end{equation}
where $\v Z_i\in\bb{R}^n$ is a Rademacher random vector and $\ln (\m A)\v Z_i$, like $\m A^{1/2}\v Z_i$ is evaluated via the Lanczos tridiagonalization algorithm (see section \ref{Lanczos methods}). Instead, we propose the following estimator, 
\begin{equation}
\widehat{\ln|\m A|} = -n\,c+2\ln\left[\frac{1}{M}\sum_{i=1}^M\exp(-n(Y_i-c)/2 )\right],
\end{equation}
where $Y_i=\ln(\v \Theta_i^\top\m A^{-1}\v \Theta_i)$, $c=\min_i Y_i$ and $\{\v \Theta_i\}_{i=1,\dots,M}$ are drawn i.i.d uniformly from the surface of the unit radius $n$-dimensional sphere centered at the origin. In this case, $\m A^{-1}\v \Theta_i$ is evaluated via the conjugate gradient method \cite{golub2013matrix}. To show that this is a valid estimator, we first note that,
\[\ln|\m A|=2\ln\int(2\pi)^{-\frac{n}{2}}\exp{(-\v z^{\top}\m A^{-1}\v z/2)}\m d\v z.\]
This equation can be derived by noting the integral of the multivariate normal density is equal to one. Thus, our goal is to produce an unbiased estimate of the integral on the RHS.

To estimate $\int \exp(-\v z^\top\m A^{-1}\v z/2)\m d \v z$ we switch to spherical coordinates. Let $\v Z\sim\c N(\mathbf{0},\m I_n)$, then, $\v Z$ can be represented as
$\v Z=R\v\Theta$ with  $\v\Theta$ being uniformly distributed on the surface of the unit radius $n$-dimensional sphere centered at the origin and $R\sim\chi_n$, independently. Then, we have the unbiased estimator,
\[
\frac{1}{(2\pi)^{n/2}}\int \exp(-\v z^\top\m A^{-1}\v z/2)\m d \v z=\bb E[(\v \Theta^\top\m A^{-1}\v \Theta)^{-n/2}]\approx\frac{1}{M}\sum_{i=1}^M(\v \Theta_i^\top\m A^{-1}\v \Theta_i)^{-n/2}.
\]
The details of this derivation are provided in the Appendix. We therefore obtain the conditional Monte Carlo estimate,
\[\widehat{\ln|\m A|}=2\ln\left[\frac{1}{M}\sum_{i=1}^M(\v \Theta_i^\top\m A^{-1}\v \Theta_i)^{-n/2}\right].\]
However, for large $n$, accurate evaluations of $(\v \Theta_i^\top\m A^{-1}\v \Theta_i)^{-n/2}$ are unlikely due to round-off errors. Instead, we let,
\[
Y_i=\ln(\v \Theta_i^\top\m A^{-1}\v \Theta_i).
\]
We then evaluate the numerically stable approximation,
\begin{align*}
\widehat{\ln|\m A|}&=2\ln\frac{1}{M}\sum_{i=1}^M \exp(-nY_i/2 )\\
&=-n\,c+2\ln\left[\frac{1}{M}\sum_{i=1}^M\exp(-n(Y_i-c)/2 )\right],
\end{align*}
where $c=\min_i Y_i$. By evaluating the terms $\exp(-n(Y_i-c)/2)$ we can alleviate further round-off errors.

\begin{figure}[H]
  \centering
  \includegraphics[width=0.7\textwidth]{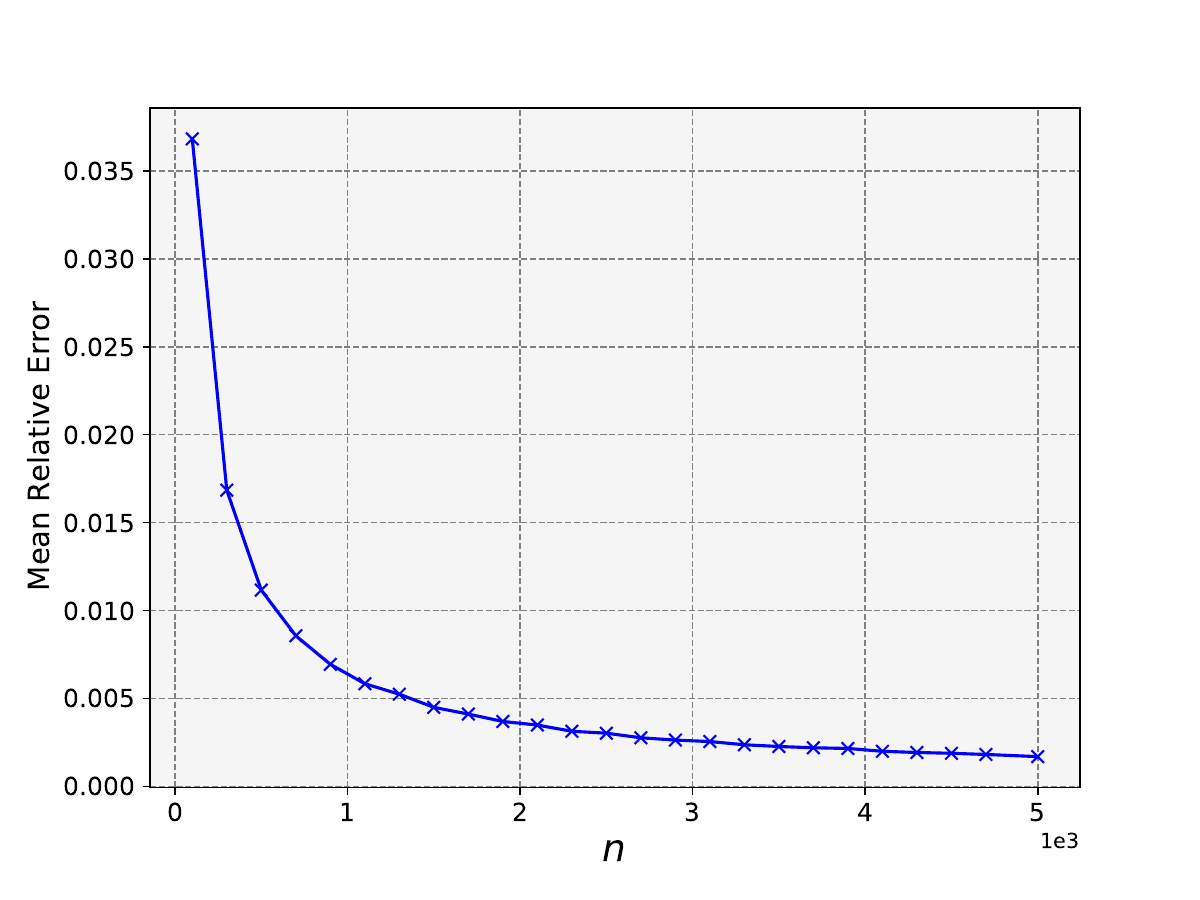}
  \caption{Mean relative error of $\widehat{\ln|\m K|}$, as $n$ increases, where $\m K$ is the squared exponential Gram matrix with $\theta_1=1$, $\theta_2=1$. Monte Carlo size: $M=10$.}
  \label{fig:f6}
\end{figure}

We note that as $\ln|\m A| = -\ln|\m A^{-1}|$, one can go about estimating $\ln|\m A^{-1}|$ and therefore evaluate the terms $Y_i=\ln(\v \Theta_i^\top\m A\v \Theta_i)$, which only require  matrix vector multiplications with $\m A$. 
As a numerical example, consider Figure \ref{fig:f6}, where we estimate the log-determinant of the matrix $\m K$, which is a squared exponential Gram matrix with $\theta_1 =1$ and $\theta_2=1$.
\section{Conclusion}
In this paper, we discuss how factorizing the PSD matrix $\m A$ can reduce variance when trying to estimate its entries using only matrix vector operations of the form $\m A\v Z$ where $\v Z$ is a random vector generated from a pre-defined distribution. We show in some examples that the unfactorized estimator can exhibit an arbitrary larger variance when compared to the factorized square root estimator. However, we also observe examples where the unfactorized estimator performs better than the factorized estimator, but the underperformance is relatively small and remains bounded from above as the size of the matrix grows. Whether this behavior of the underperformance holds true in general for all matrices $\m A$ is an open question.

Furthermore, we show how factorization can achieve variance reduction when estimating $\tr(\m A\m W)$, where $\m A\in\bb{R}^{n\times n}$ is a PSD matrix and $\m W \in \bb{R}^{n \times n}$ is a symmetric matrix . Similar to arbitrary matrix entry estimation, we observe numerical examples where the factorized estimator significantly outperforms the unfactorized estimator. However, when $\v Z$ is Rademacher, this outperformance is not theoretically guaranteed for all choices of $\m A$ and $\m W$. Ideally, we'd like a theoretical framework for identifying when the factorized estimator will work most in our favor.

Finally, we provide a conditional Monte Carlo estimate for the log-determinant of a symmetric positive-definite matrix $\m A$ by treating it as the normalizing constant of a probability density. We note that it is possible to implement Markov chain Monte Carlo (MCMC) techniques such as particle filters to provide better estimates for the quantity $\bb{E}\left[(\v \Theta^\top\m A^{-1}\v \Theta)^{-n/2}\right]$.

\clearpage
\appendix
\section{Proof of Theorem \ref{min_gauss}}
The following proof is not explicitly stated in
\cite{martens2012estimating} and so we include it here for completeness. 
\begin{proof}
Since $\m A=\m B\m C^\top$, we have $\v b_{i,:}^\top\v c_{i,:}=a_{i,i}$.
Applying the Cauchy--Schwarz inequality to \eqref{bekas_var} gives
\[
\begin{aligned}
\var_G[\hat a_{i,i}]
&=\|\v b_{i,:}\|^2\|\v c_{i,:}\|^2+a_{i,i}^2\\
&\geq (\v b_{i,:}^\top\v c_{i,:})^2+a_{i,i}^2
=2a_{i,i}^2.
\end{aligned}
\]
If $\m C=\m B$ and $\m B\m B^\top=\m A$, then
$\|\v b_{i,:}\|^2=a_{i,i}$, so
$\var_G[\hat a_{i,i}]=2a_{i,i}^2$.
Thus any factorization $\m A=\m B\m B^\top$ attains the minimum Gaussian
variance for every diagonal entry.
\par
\end{proof}

\section{Uniform Spherical Estimator}
\subsection{Proof of Theorem \ref{sphere_element_var}}
\begin{proof}
Suppose $\v Z = (Z_1,\dots,Z_n)^{\top}$ is uniformly distributed on the surface of the unit radius $n$-dimensional sphere centered at the origin. Using spherical coordinates, it is not difficult to show that the vector $\v X=\v Z\odot\v Z$ follows a Dirichlet distribution  over the simplex:
\[
f(\v x)=\frac{\Gamma(n/2)}{\pi^{n/2}}\prod_{i=1}^n x_i^{-1/2},\quad x_i\in[0,1],\quad \sum_{i=1}^nx_i=1.
\]
Therefore, using the well-known mean and variance formula for the Dirichlet distribution we have,
\begin{equation}
\bb E[\v X]=\v 1/n,\quad \var[\v X]=\frac{1}{n(n/2+1)}\left[\m I_n-\frac{1}{n}\v 1\v 1^\top\right].
\end{equation}
Using spherical coordinates, we also know that $\bb{E}[Z_iZ_j]=0$ for $i\neq j$. Therefore,
\[\bb{E}[\v Z\v Z^{\top}]=\frac{1}{n}\m I_n,\]
and, 
\[\bb{E}\left[n\left(\m B \v Z\right)\left(\m C \v Z\right)^{\top}\right] =n\bb{E}\left[\m B\v Z \v Z^{\top}\m C^\top\right]= \m A.\]
To prove the variance, we first note that,
\[
\bb{E}[(\hat{a}_{i,j}/n)^2]=\bb{E}[(\left[\m B \v Z\right]_i\times\left[\m C \v Z\right]_j)^2] = \sum_{k,l,m,n}b_{i,k}c_{j,l}b_{i,m}c_{j,n}\bb{E}[Z_{k} Z_{l} Z_{m} Z_{n}].
\]
Representing $\v Z$ in spherical coordinates, we obtain the formula,
\[\bb{E}[Z_{k} Z_{l} Z_{m} Z_{n}] = c_1[\delta_{kl}\delta_{mn}+\delta_{km}\delta_{ln}+\delta_{kn}\delta_{lm}]+(c_2-3c_1)[\delta_{kl}\delta_{lm}\delta_{mn}].\]
The constants $c_1$ and $c_2$ are given by,
\begin{align*}
c_1 &= \var[\v X]_{p,q}+ \bb E[\v X]_p^2=  \frac{1}{n(n+2)},\qquad p \neq q,\qquad \text{and}\\
c_2 &=\var[\v X]_{p,p}+\bb E[\v X]_p^2= \frac{3}{n(n+2)}.
\end{align*}
Thus $c_2-3c_1=0$, and,
\begin{align*}
\bb{E}[(\hat a_{i,j}/n)^2]
                &= c_1\sum_{k,l,m,n}b_{ik}c_{jl}b_{im}c_{jn}[\delta_{kl}\delta_{mn}+\delta_{km}\delta_{ln}+\delta_{kn}\delta_{lm}]\\
                &= c_1\left[(\v b_{i,:}^{\top}\v c_{j,:})^2 +(\v b_{i,:}^{\top}\v b_{i,:})(\v c_{j,:}^{\top}\v c_{j,:})+(\v b_{i,:}^{\top}\v c_{j,:})^2\right]\\
                &= c_1\left[2a_{i,j}^2 +\|\v b_{i,:}\|^2\|\v c_{j,:}\|^2\right].
\end{align*}
Therefore,
\begin{align*}
\var_{S}[(\hat{a}_{i,j}/n)] &= \bb{E}\left[(\hat{a}_{i,j}/n)^2\right]-a^2_{i,j}/n^2\\ 
                &= (2c_1-1/n^2)a_{i,j}^2+c_{1}\|\v b_{i,:}\|^2\|\v c_{j,:}\|^2.\\ 
\end{align*}
Hence, the variance for $\hat{a}_{i,j}$ is,
\begin{align*}
\var_{S}[(\hat{a}_{i,j})]&= n^2\var_{S}[(\hat{a}_{i,j}/n)]\\
&= \frac{n-2}{n+2}a_{i,j}^2+\frac{n}{n+2}\|\v b_{i,:}\|^2\|\v c_{j,:}\|^2.
\end{align*}
\end{proof}

\subsection{Proof of Theorem \ref{sphere_quad_var}}
\begin{proof}
Suppose $\m\Sigma'$ has eigenvalues $\{\lambda_i\}$, and orthonormal eigendecomposition  $\m Q\m \Lambda \m Q^{\top}$. Then, as the random variable that is distributed uniformly on the surface of the $n$-dimensional sphere centered at the origin is invariant to orthogonal rotations, we obtain,
\[
\var[\v Z^\top\m\Sigma'\v Z]=\var[\v Z^\top\m\Lambda\v Z]=\var[\v\lambda^\top\v X]=\v\lambda^\top\var[\v X]\v\lambda.
\]
We notice that $\tr((\m\Sigma')^\top\m\Sigma')=\v\lambda^\top\v\lambda$. Therefore,
\[\var[n\v Z^\top\m\Sigma'\v Z]=\frac{n}{n+2}\times2\left(\tr(\m\Sigma'^\top\m\Sigma')-\frac{(\sum_i\m\Sigma'_{i,i})^2}{n}\right).\]
To see that this gives the variance for $n\v Z^{\top}\m\Sigma\v Z$, we note,
\[\v Z^{\top}\m\Sigma'\v Z = \v Z^{\top}\left[(\m\Sigma + \m\Sigma^{\top})/2\right]\v Z=\v Z^{\top}\m\Sigma\v Z.\]
\end{proof}
We note that when $\v Z$ is distributed uniformly on the unit radius complex sphere instead, the variance formula is given in Tropp \cite{tropp2020randomized}.
\clearpage
\section{Log-Determinant }
Suppose $\m\Sigma\in\bb{R}^{n\times n}$ is a symmetric positive-definite matrix. Let $\v Z\sim\c N(\mathbf{0},\m I_n)$. Then, $\v Z$ can be represented as
$\v Z=R\v\Theta$ with  $\v\Theta$ being uniformly distributed on the surface of the unit radius $n$-dimensional sphere centered at the origin and $R\sim\chi_n$, independently. Then using the standard normal as a change of measure we get the following,
\[
\begin{split}
\frac{1}{(2\pi)^{n/2}}\int \exp(-\v z^\top\m\Sigma^{-1}\v z/2)\m d \v z&=\bb E \left[\exp(-\v Z^\top\m\Sigma^{-1}\v Z/2+\|\v Z\|^2/2)\right]\\
&=\bb E \left[\exp(-R^2\v \Theta^\top\m\Sigma^{-1}\v \Theta/2+R^2/2)\right]\\
&=\frac{1}{2^{n/2}\Gamma(n/2)}\bb E \left[\int_0^\infty r^{n/2-1}\exp(-r\v \Theta^\top\m\Sigma^{-1}\v \Theta/2)\m d r\right]\\
&=\bb E \left[ (\v \Theta^\top\m\Sigma^{-1}\v \Theta)^{-n/2} \right]\\
&\approx\frac{1}{M}\sum_{i=1}^M(\v \Theta_i^\top\m\Sigma^{-1}\v \Theta_i)^{-n/2}.
\end{split}
\]
We use the following integral formula to evaluate the integral on line 3, 
\[
\int_0^\infty r^{n/2-1}\exp(-r\alpha/2)\m d r=\alpha^{-n/2}2^{n/2}\Gamma(n/2).
\]
Thus a conditional Monte Carlo estimate for the log-determinant is, \[\widehat{\ln|\m\Sigma|}=2\ln\left[\frac{1}{M}\sum_{i=1}^M(\v \Theta_i^\top\m\Sigma^{-1}\v \Theta_i)^{-n/2}\right].\]

\bibliographystyle{plainnat}
\bibliography{bibliography}

\end{document}